\documentclass[runningheads]{llncs}
\usepackage[T1]{fontenc}
\usepackage{graphicx}
\usepackage{subfig}
\usepackage{float}
\usepackage{caption}
\begin{document}
\title{Automatic Signboard Recognition in Low Quality Night Images}
\author{Manas Kagde\inst{1} \and
Priyanka Choudhary\inst{1}\orcidID{0009-0003-7958-6402} \and
Rishi Joshi\inst{2} \and
Somnath Dey\inst{1}\orcidID{0000-0002-5566-1944}}
\authorrunning{M. Kagde et al.}
\institute{Indian Institute of Technology Indore, Indore, Madhya Pradesh, India \and
Shri Vaishnav Vidyapeeth Vishwavidyalaya, Indore, Madhya Pradesh, India
\email{}\\
%\url{http://www.springer.com/gp/computer-science/lncs} 
% \and
% \\
% \and
% Computer Science and Engineering, Indian Institute of Technology Indore, Khandwa Rd, Simrol, 453552, Madhya Pradesh, India
% \email{}
}
\maketitle              % typeset the header of the contribution
\begin{abstract}
An essential requirement for driver assistance systems and autonomous driving technology is implementing a robust system for detecting and recognizing traffic signs. This system enables the vehicle to autonomously analyze the environment and make appropriate decisions regarding its movement, even when operating at higher frame rates. However, traffic sign images captured in inadequate lighting and adverse weather conditions are poorly visible, blurred, faded, and damaged. Consequently, the recognition of traffic signs in such circumstances becomes inherently difficult. This paper addressed the challenges of recognizing traffic signs from images captured in low light, noise, and blurriness. To achieve this goal, a two-step methodology has been employed. The first step involves enhancing traffic sign images by applying a modified MIRNet model and producing enhanced images. In the second step, the Yolov4 model recognizes the traffic signs in an unconstrained environment. The proposed method has achieved 5.40\% increment in mAP@0.5 for low quality images on Yolov4. The overall mAP@0.5 of 96.75\% has been achieved on the GTSRB dataset. It has also attained mAP@0.5 of 100\% on the GTSDB dataset for the broad categories, comparable with the state-of-the-art work.
\end{abstract}
\section{Introduction}
\begin{figure}
\vspace{-6mm}
\begin{minipage}{.28\textwidth}
  \centering
  \includegraphics[width=2cm]{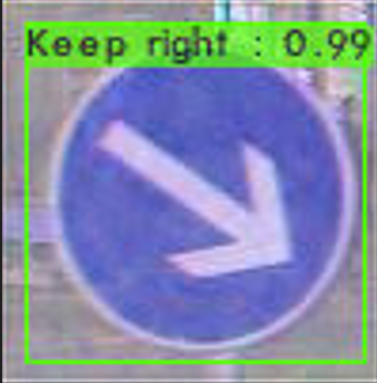}
   % \centering
  \captionof{figure}{Keep Right traffic signboard with predicted bounding boxes}
  \label{fig:traffic_sign_predicted}
\end{minipage}
% \vspace{-6mm}
% \vspace{-6mm}
\hfill
% \centering
\begin{minipage}{.28\textwidth}
  \centering
  \includegraphics[width=2cm]{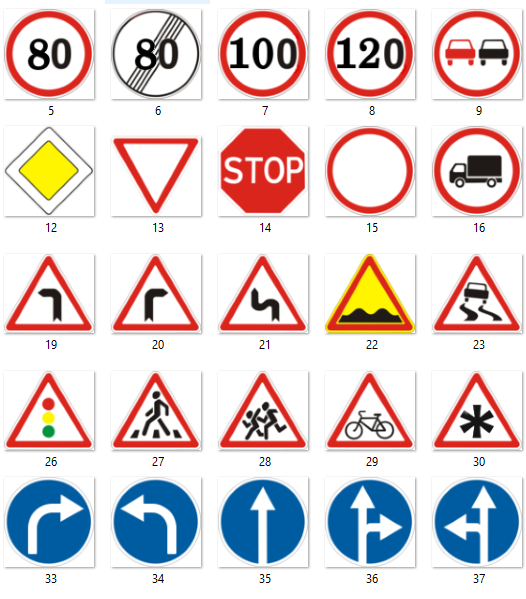}
   \centering
  \captionof{figure}{Sample GTSRB Traffic Sign Images}
  \label{fig:traffic_sign_broad}
\end{minipage}%
% \hspace{10mm}
\hfill
% \centering
\begin{minipage}{.34\textwidth}
  \centering
  \includegraphics[width=4cm]{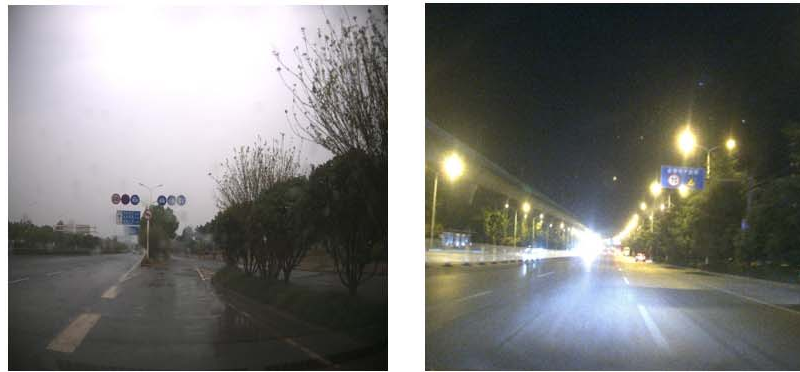}
   \centering
  \captionof{figure}{unfavorable environmental conditions on road}
  \label{fig:night}
\end{minipage}
\vspace{-5mm}
\end{figure}

%\subsection{A Subsection Sample}
Traffic sign recognition is a technology by which intelligent systems like Advanced Driver Assistance Systems (ADAS)~\cite{1} can recognize the traffic sign boards on the roads, e.g., ‘Pedestrians’ or ‘Turn Left,’ etc. ADAS are intelligent machines that increase the vehicle’s safety by guiding the driver about front road conditions and providing better driving instructions. ADAS and vehicular automation rely on traffic sign recognition algorithms for roadway decision-making processes. Traffic sign detection encompasses the tasks of generating bounding boxes and classifying the content within those bounding boxes. Fig.~\ref{fig:traffic_sign_predicted} shows the predicted bounding box for the ‘Keep right’ traffic sign. Traffic sign recognition can be done on both broader and fine-grained categories. In the fine-grained classes, each traffic sign acts as a single category, whereas in the case of broader categories, all similar kinds of traffic signs are grouped into one class. For example, in Fig.~\ref{fig:traffic_sign_broad}, all circular traffic signs having background color white and border color red can be grouped under ‘Prohibitory’ signs (80kmph, 100kmph, 120kmph, etc.). In contrast, all the triangular traffic signs with white color in the background and red border color are grouped as ‘Danger’ signs (Yield, Turn Left, Turn Right, etc.). Traffic sign recognition becomes even more challenging for fine-grained classes with unfavorable environmental and weather conditions (Fig.~\ref{fig:night})~\cite{27} such as low contrast, noise, blur, faded, damaged, and occluded due to trees, persons or objects.
\vspace{-2mm}

Traffic sign recognition employs the traditional approach and the state-of-the-art methodology. Conventional approaches like color segmentation~\cite{9}, color-shape change~\cite{10} and Histogram of Oriented Gradients (HOG)~\cite{11,1219} leverages color, shape, and gradient features to extract the information from images. However, these approaches face challenges due to illumination changes. Modern state-of-the-art algorithms apply Convolutional Neural Network (CNN)-based models~\cite{13,14,15} with different data preprocessing methods to reduce the noise in an image. Current state-of-the-art detectors like Faster R-CNN~\cite{3,6}, Single Shot Detector (SSD)~\cite{5,7}, and Yolov3~\cite{4,8} have recently gained popularity due to notable advancements in performance on object detection and remarkable speed capabilities. Nevertheless, they exhibit computational complexity and encounter issues related to over-fitting as the network size expands. Additionally, their predictive accuracy diminishes when dealing with fine-grained categories in adverse environmental and weather conditions.

Our proposed work adopted a two-step methodology for traffic sign recognition. Initially, low contrast, noisy, and blurry images of traffic signs are enhanced by a modified MIRNet~\cite{18} model. Then, the enhanced images are passed further to Yolov4~\cite{17} for prediction. The major contribution of the proposed approach is that it can produce better results on the fine-grained traffic sign categories, particularly when dealing with low quality images. 

The rest of this article is structured as follows: Section~\ref{sec:two} presents the related work. Section~\ref{sec:three} presents our proposed approach. Subsequently, the experimental evaluation and its results are presented in Section~\ref{sec:four}, and Section~\ref{sec:five} offers our conclusions.

\section{Related Work}
\label{sec:two}
Extensive research is being conducted in the field of traffic sign recognition, encompassing two primary approaches: traditional methodologies and modern state-of-the-art techniques. Traditional approaches~\cite{9,10,11} include algorithms like Histogram of oriented gradients~\cite{1219}, Color-Shape change~\cite{10} and color segmentation~\cite{9} based on traffic sign recognition. Modern state-of-the-art algorithms make use of CNN~\cite{13,14,20}, Faster R-CNN~\cite{3,6} and Yolov4~\cite{16,17} based approaches.

Sun et al.~\cite{13} utilized hough transform and image segmentation to localize traffic signs in the  German Traffic Sign Recognition Benchmark (GTSRB)~\cite{2} dataset for circular-shaped signs. They used deep learning to detect traffic signs on the roads. The model has achieved an impressive accuracy of 98.2\%. Notably, the authors simplified the problem by categorizing the traffic signs into two broader groups instead of the original 43 classes. 

Patil et al.~\cite{14} used GTSRB~\cite{2} dataset to train the CNN model with eight layers. They employed ReLU and Softmax to lessen the computational burden, number of dimensions, and overfitting problems generated by deep fully connected convolutional layers. The study achieved an accuracy of 95\%. Han et al.~\cite{15} proposed a traffic sign recognition method based on LeNet architecture. Shape and color segmentation techniques were employed to extract Regions of Interest (ROI), potentially containing traffic signs. The proposed methodology demonstrated a remarkable accuracy of 96.2\% when evaluated on the test images from the GTSRB~\cite{2}.

Khnissi et al.~\cite{16} upgraded Yolov4 with a new Yolov4-tiny-based compact classifier. The model could recognize all 43 GTSRB~\cite{2} traffic signs with an average accuracy of 95.44\% while trying to save computational power and processing time by squeezing the network. Novak et al.~\cite{4} proposed a model for road object detection into five broad classes. The method achieved a classification accuracy of 99.2\% for detected traffic signs in different weather conditions for their independently created dataset. Jianjun et al.~\cite{5} used SSD~\cite{7}, which integrates the Path Aggregation Network (PAN) in conjunction with the Receptive Field (RF) module, yielding enhanced accuracy in the results. The authors also claimed that as compared with the common object detection algorithms such as Faster R-CNN~\cite{21}, RetinaNet, and Yolov3~\cite{22}, the SSD-RP could achieve a better balance between detection time and detection precision.

% \vspace{-3pt}
\section{Proposed Method}
\label{sec:three}
% \vspace{-3mm}
This paper introduced a novel two-step methodology for traffic sign recognition for various challenging conditions. In the initial step, a Modified MIRNet~\cite{18} was employed to enhance the quality of low quality test images. Subsequently, the Yolov4~\cite{17} model was utilized in the second step to predict the improved output results in terms of mean Average Precision (mAP) and accuracy. The spatial attention layer of the MIRNet~\cite{18} model was modified and trained the model to generate better-enhanced test image results. 
% The GTSRB~\cite{2} test images that are noisy, blurry, shadowy, and low contrast are enhanced through the modified MIRNet~\cite{18} model. Further, these enhanced images were provided to Yolov4~\cite{17} for prediction. 
% We are able to generate better mAP and accuracy for both the fine-grained GTSRB~\cite{2} 43 classes and GTSDB~\cite{1219} has 4 broad categories. We produced better mAP on the noisy images as well.

\begin{figure}
\vspace{-4mm}
\centering
\includegraphics[width=0.8\textwidth]{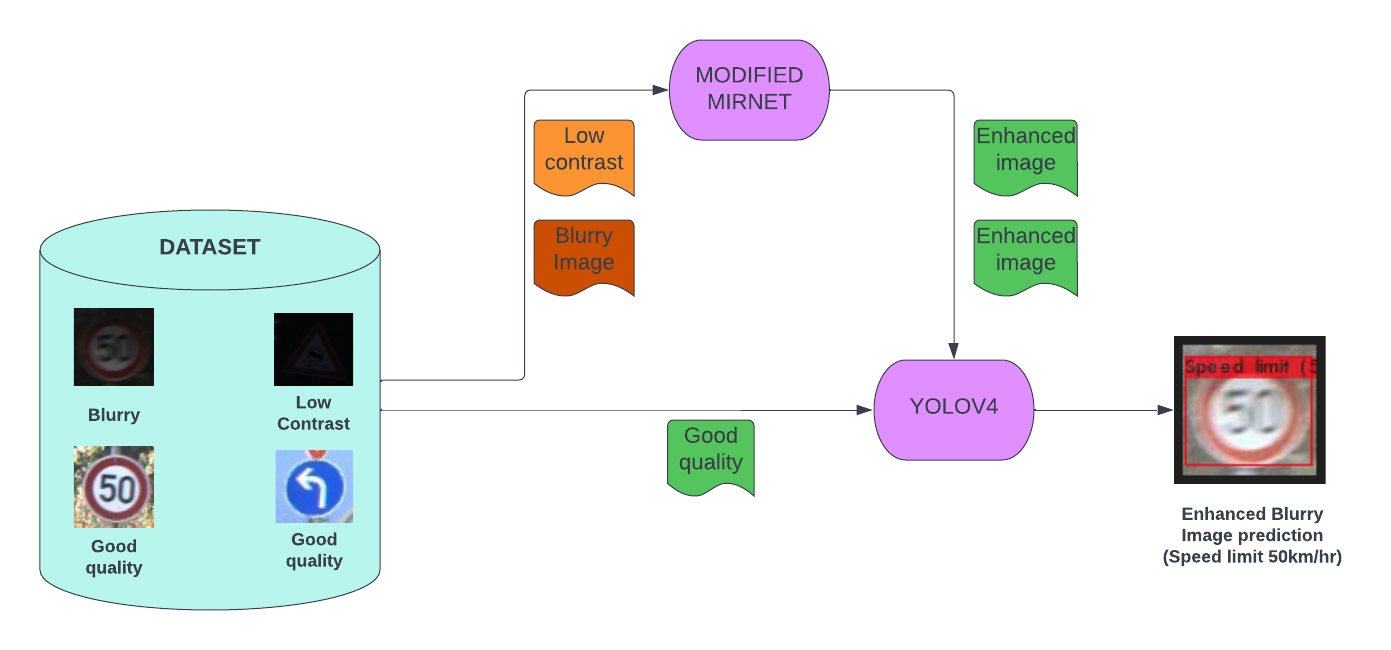}
\caption{Flow diagram of the proposed architecture}
\label{proposed}
\vspace{-3mm}
\end{figure}

Fig.~\ref{proposed} illustrates the flow diagram of the proposed architecture. The two-step process is mentioned in the proposed approach, wherein the first step, the low quality images are passed to the Modified MIRNet model. Subsequently, in the second step, the enhanced images obtained from the Modified MIRNet model are fed into the Yolov4 model for prediction. Moreover, the good-quality images are fed directly to the Yolov4 model for prediction.
\vspace{-3mm}

\subsection{Original Architecture of Yolov4}
The Yolov4 model has been used as our base model for prediction. Regarding optimal speed and accuracy, Yolov4~\cite{17} performed much better at higher frame rates. Yolov4~\cite{17} comes in one stage detector category, where detection and recognition happen in a single shot during training. 
% Fig.~\ref{yolov4} shows the architecture of the Yolov4 model.
% \begin{figure}
% \vspace{-3mm}
% \centering
% \includegraphics[width=\textwidth]{pictures/architecture.PNG}
% \caption{Architecture of Yolov4 ~\cite{17}}
% \label{yolov4}
% \vspace{-3mm}
% \end{figure}
There are the following building blocks of Yolov4:
\vspace{-3mm}
\subsubsection{Backbone}
It generates a feature map with the help of CSPDarknet53~\cite{23}, like a DenseNet architecture.
\vspace{-3mm}
\subsubsection{Neck} 
While using Spatial Pyramid Pooling (SPP)~\cite{24}, Path Aggregation Network (PAN)~\cite{25}, it aggregates features from different backbone stages with several top-down and bottom-up paths to segregate the useful contextual features.
\vspace{-6mm}
\subsubsection{Head} 
Yolov4 uses Yolov3~\cite{8} in the head portion, responsible for the classification and the bounding box generation. Despite these, Yolov4 introduced new features, Bag of Freebies (BoF) and Bag of Specials (BoS), to leverage a combination of cost-effective modifications in training strategy and deliver significant accuracy improvements, respectively.
\subsection{Modified MIRNet}
\begin{figure}
\vspace{-3mm}
\centering
\subfloat[MIRNet~\cite{18} Architecture]{\includegraphics[width=\textwidth]{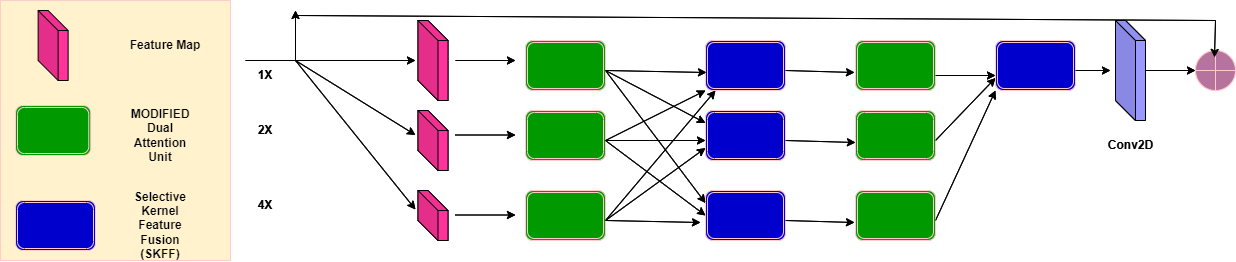}
\label{fig:mirnet_a}}
\\[12pt]
\subfloat[Modified dual attention unit of MIRNet Architecture]{\includegraphics[width=\textwidth]{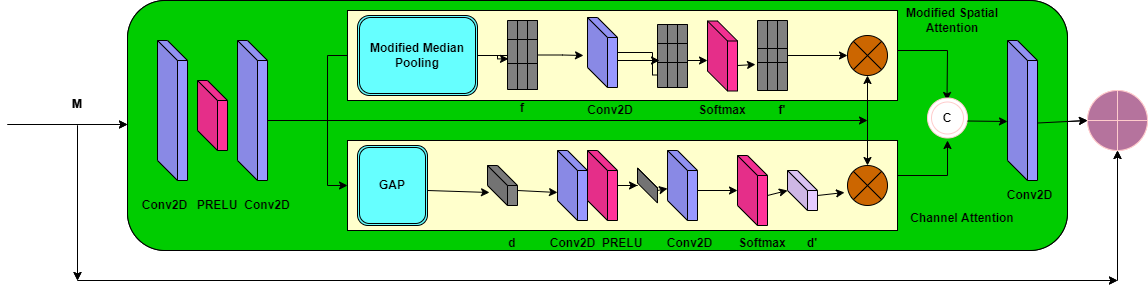}
\label{fig:mirnet_b}}
\caption{Proposed Modified MIRNet Architecture Diagram}
\label{fig:modified_mir}
\vspace{-6mm}
\end{figure}

Image enhancement is a sophisticated methodology to generate a refined and superior-quality image by utilizing low quality source images. To accomplish this task, the most suitable approach involves utilizing a CNN-based model known as MIRNet~\cite{18}. It has proven to be highly effective in generating solutions for this purpose. This study has modified the architecture of MIRNet, shown in Fig.~\ref{fig:modified_mir}. The different resolution low contrast, noisy, and blurred images are passed through the modified MIRNet model. The feature maps of each image with the different resolutions are then inputted into the modified Dual Attention layer for feature extraction as shown in Fig.~\ref{fig:mirnet_b}. These are then passed to the Selective Kernel Feature Fusion (SKFF) for feature aggregation. The Dual Attention Unit and SKFF are applied again to produce the final results.
% Here, the Global Average Pooling + Global Max pooling combinations in the spatial attention layer of the MIRNet model are replaced by the Median pooling. 
% The different resolution low contrast, noisy, and blur images are fed to the modified MIRNet model. The feature maps of each image with the different resolutions are passed into the modified Dual Attention layer for feature extraction as shown in Fig.~\ref{fig:mirnet_b}. These are then passed to the Selective kernel feature fusion (SKFF) for feature aggregation. The dual attention unit and SKFF are applied again to produce the final results.
% \vspace{-10mm}
\subsubsection{Multi-Scale Residual Block (MRB)}
The primary component encompasses the Selective Kernel Feature Fusion, a modified Dual Attention block, and an upsampling-downsampling module. Fig.~\ref{fig:mirnet_a} shows the MRB’s structure in MIRNet~\cite{18}. MRB facilitates the sharing of valuable information among interconnected layers operating in parallel. This process leads to the generation of improved-quality features from initially low quality images.
\vspace{-2mm}
\subsubsection{Selective Kernel Feature Fusion (SKFF)}
The feature aggregation network employs a self-attention mechanism to integrate features obtained from various resolutions. It is responsible for the effective fusion and representation of information. Fig.~\ref{fig:mirnet_a} shows the SKFF structure in MIRNet~\cite{18}.
\vspace{-3mm}
\subsubsection{Dual Attention Unit (DAU)}
The network incorporates a feature extraction mechanism that effectively suppresses irrelevant features and selectively allows only highly relevant features to be retained. Fig.~\ref{fig:mirnet_a} shows the dual attention mechanism in MIRNet~\cite{18}. The term "dual" refers to utilizing both spatial and channel attention to capture valuable features. The spatial channel combines Global Average Pooling (GAP) and Global Max Pooling (GMP) techniques to capture the inter-spatial dependencies in the convolutional features. On the other hand, channel attention focuses on extracting connections between different channels within the feature maps. This is achieved through GAP to encode global context across spatial dimensions and two convolutional layers with a sigmoid activation function.
%
% \vspace{-3mm}
% \paragraph{Modified Spatial Attention}
\subsubsection{Modified Spatial Attention}
Fig.~\ref{fig:mirnet_b} shows the modified dual attention mechanism in MIRNet~\cite{18}. The GAP+GMP combination in the Spatial Attention layer has been substituted with Median Pooling. Average pooling is a technique that disperses noise across all blocks within a given context. In contrast, median pooling is employed to eliminate noise from the data. It also facilitates elucidating edges that enable extracting highly valuable spatial feature maps. Additionally, it aids in the consolidation of two distinct operations that are conventionally executed individually on each feature map.
\vspace{-3mm}
\section{Experimental Results and Analysis}
\label{sec:four}
This section highlights the dataset, data preparation, training, and results of our proposed approach. The comparison with existing approaches is also presented in this section.
\vspace{-2mm}
\subsection{Datasets and Preprocessing}
The GTSDB~\cite{1219} and GTSRB~\cite{2} datasets were utilized for our experiment since they contain a wide variety of traffic signs which help to make a fine-grained classification. Table~\ref{tab:dataset} shows the categorization of the GTSDB~\cite{1219} and GTSRB~\cite{2} datasets, respectively. The GTSDB dataset contains lesser training and testing images than the GTSRB dataset. The training images of the GTSDB~\cite{1219} contain full-size images of the traffic sign on roads whereas GTSRB~\cite{2} contains cropped smaller-size traffic sign images. The GTSDB dataset has four major classes: Prohibitory, Mandatory, Danger, and Other. In contrast, the GTSRB dataset encompasses a more detailed classification framework with 43 fine-grained classes.

\begin{table}
\vspace{-5mm}
% \vspace*{12pt}
\begin{center}
\caption{Details of GTSDB~\cite{1219} and GTSRB~\cite{2} dataset.}
\begin{tabular}{ |c|c|c|c|c|c|  }
\hline
 \textbf{Dataset} &\textbf{Total Images}& \textbf{Training Images} & \textbf{Test Images} &\textbf{Classes} & \textbf{Categories}\\
 \hline
 \hline
 GTSDB~\cite{1219} & 900 & 600 & 300 & 4 & 4\\
 \hline
 \hline
 GTSRB~\cite{2} & 51838 & 39209 & 12629 & 43 & 43\\
 \hline
 \hline
\end{tabular}
\label{tab:dataset}
\end{center}
\vspace{-5mm}
\end{table}
In order to effectively train the network, it is necessary to preprocess and convert the images and annotations from the dataset into the desired format. Specifically, the conversion of images from ppm and png formats into jpg format. Additionally, annotation files in CSV format for each image are separated in both the GTSDB and GTSRB datasets, resulting in input for the Yolov4 model.
\vspace{-2mm}
\subsection{Training and Validation on Modified MIRNet}
The main objective of the proposed approach is to enhance the low quality traffic images using the modified MIRNet~\cite{18} model. The training of the Modified MIRNet model requires both low and high-quality images of the same training image. However, GTSRB and GTSDB datasets do not contain a pair of low and high-quality images of the same input image. So, the modified MIRNet model was trained with the Low Light (LOL)~\cite{26} benchmark dataset, which contains low and high-quality general-purpose image pairs. Specifically, 300 pairs for training and 185 pairs for validation were utilized in our study.

The parameters used for training were \textit{max-batches} as 4, \textit{image-size} of 600 × 400, \textit{random-crops} of 128 × 128. The parameters for checking the performance of the Modified MIRNet model were: \textit{Loss} and \textit{PSNR} (Peak Signal Noise Ratio). \textit{Loss} tells how close is our predicted image from the actual image, and \textit{PSNR} gives the ratio of the image signal to that of the noise present in it. So, the lesser the noise, the higher \textit{PSNR} yield better prediction results.

\begin{figure}
\vspace{-2mm}
\centering
\begin{minipage}{5.5cm}
  \centering
  \includegraphics[width=5.5cm]{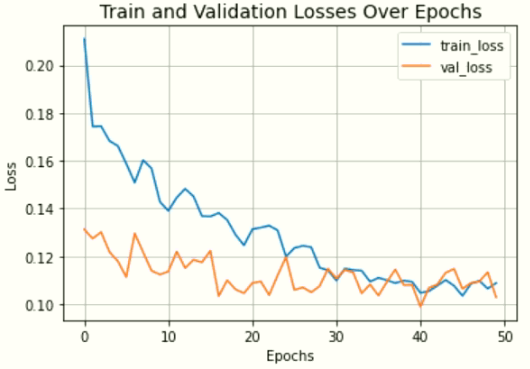}
   \centering
  \captionof{figure}{Epochs vs Loss curve with Modified MIRNet~\cite{18}}
  \label{fig:epoc_loss}
\end{minipage}%
% \hspace{3mm}
\hfill
\begin{minipage}{5.5cm}
  \centering
  \includegraphics[width=5.5cm]{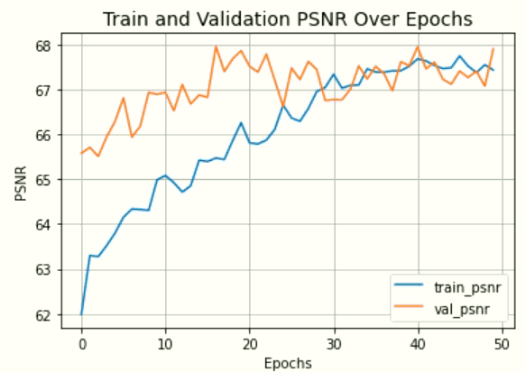}
   \centering
  \captionof{figure}{Epochs vs PSNR curve with Modified MIRNet~\cite{18}}
  \label{fig:epoc_psnr}
\end{minipage}
\end{figure}

The \textit{Epochs} vs \textit{Loss} curve for the training and validation is shown in Fig.~\ref{fig:epoc_loss}. It has been observed that the validation loss is within the range of 0.10 to 0.12 after 20\textsuperscript{th} iteration and reaches the lowest at 0.10 at the 40\textsuperscript{th} iteration. The \textit{Epochs} vs. \textit{PSNR} curve, as shown in Fig.~\ref{fig:epoc_psnr} indicates that the validation \textit{PSNR} is always in the ratio of 67\% to 68\% after 30\textsuperscript{th} iterations and reached the maximum at the 40\textsuperscript{th} iteration.
\vspace{-7mm}
\subsection{Training and Validation on Yolov4 Model}
The Yolov4 model was trained on both GTSRB~\cite{2} and GTSDB~\cite{1219} datasets to test the performance of traffic sign recognition on broad categories and fine categories, respectively. The training parameters used in the model for the GTSRB dataset were: \textit{batch} = 64, \textit{subdivisions} = 32, \textit{saturation} = 1.5, \textit{exposure} = 1.5, \textit{hue} = 0.1, \textit{learning-rate} = 0.01, \textit{max-batches} = 86000, \textit{steps} = [68800, 77400], \textit{scales} = [0.1, 0.1] and \textit{classes} = 43.

On the other hand, the training parameters used in the GTSDB dataset were \textit{max-batches} = 8000, \textit{steps} = [6400, 7200], \textit{exposure} = 1.5 and \textit{hue} = 0.1. Here, classes are modified to 4 and each of its preceding convolutional filters to 27. The training took around 5000 iterations to get better results.

The performance of the Yolov4 model is usually measured with the help of precision, recall, \textit{f1\textendash score} and Mean Average Precision \textit{(mAP)} of the training. \textit{Precision} can be defined as the correctness of the model prediction compared to all the model predictions. \textit{Recall} can be defined as the correctness of the model predictions compared to the actual. \textit{f1\textendash score} is the harmonic mean of the \textit{precision} and the \textit{recall} values, and \textit{mAP} is the mean of the average precision of each class.
% We have achieved the \textit{precision} = 99\%, \textit{recall} = 100\%, \textit{f1\textendash score} = 99\%, and \textit{mAP@0.5} = 99.88\%, during the training of Yolov4~\cite{17} model with GTSRB~\cite{2} dataset. Whereas, we have achieved the \textit{precision} = 100\%, \textit{recall} = 100\%, \textit{f1\textendash score} = 100\%, and \textit{mAP@0.5} = 99.89\%, during the training of Yolov4~\cite{17} model with GTSDB~\cite{1219} dataset.
\subsection{Results of Proposed Model}
In our proposed approach, low quality images like blurry, noisy and low contrast were passed through the modified MIRNet model as mentioned in Fig.~\ref{proposed}. In contrast, the good quality images were directly fed to the Yolov4 model for prediction.
\subsubsection{Results on Good-Quality Images}
\begin{figure}
\vspace{-3mm}
\subfloat{\includegraphics[width=6cm]{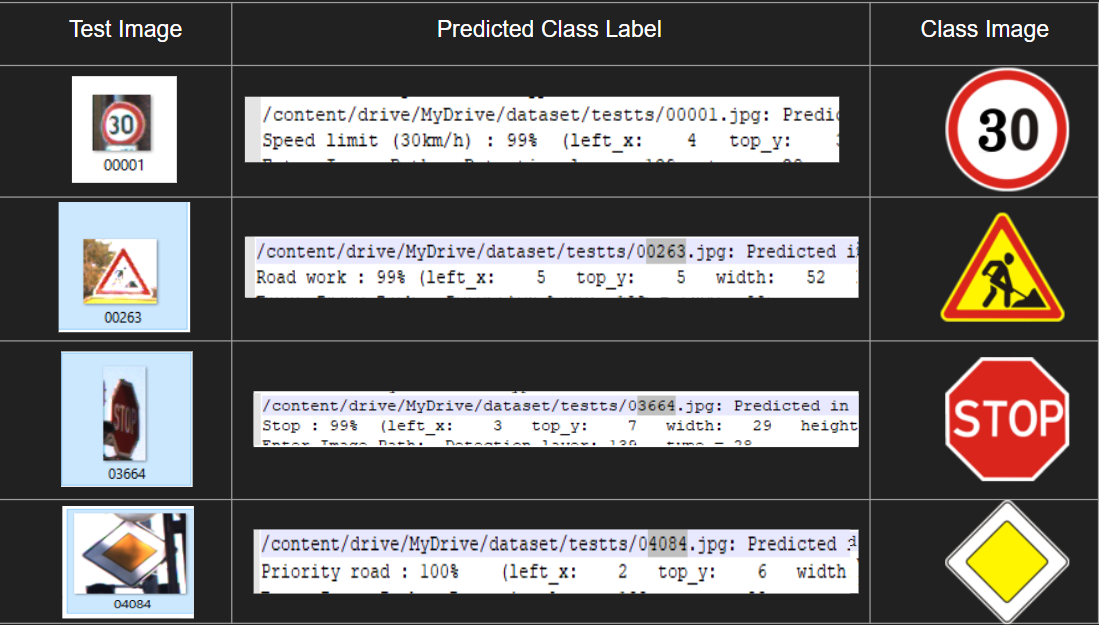}
\centering
\label{fig:13a}}
% \hfill
\subfloat{\includegraphics[width=6cm]{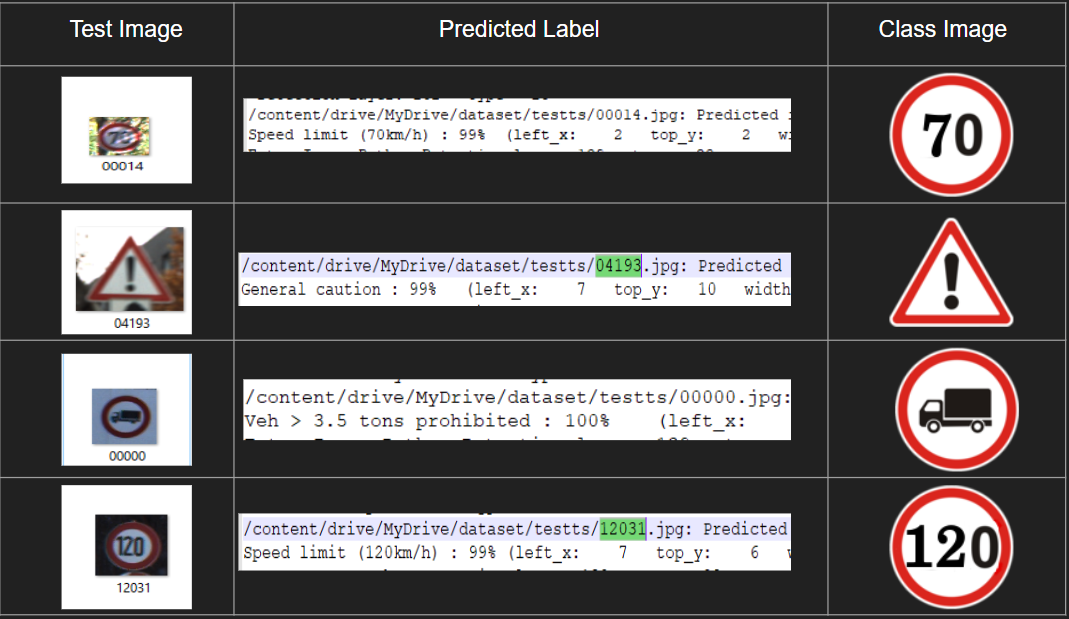}
\label{fig:13b}}
\caption{Prediction results on the good quality GTSRB~\cite{2} test images}
\label{fig:13}
\vspace{-7mm}
\end{figure}

The good quality images in the GTSRB~\cite{2} test set were passed to the Yolov4 model for the prediction. Fig.~\ref{fig:13} shows that the proposed model correctly predicts the traffic sign in good quality test images with a probability of more than 99\% for the GTSRB dataset. Hence, the proposed model performed well on the good quality test images.
% GTSDB dataset~\cite{1219} contains good quality images of the traffic signs from various the roads and size of the captured images is larger that GTRSB images. The traffic signs inside these large sized images are cropped. These cropped images are further passed to our Yolov4 model for prediction. We are able to predict all GTSDB images with more than 90\% probability correctly. %Fig.~\ref{fig:14} shows sample results of cropped images of GTSDB~\cite{1219} dataset.
% \begin{figure}
%     \centering
%     \includegraphics[width=8cm]{pictures/GTSRB_GTSDB_25200.PNG}
%     \caption{Yolov4~\cite{17} prediction results on GTSDB cropped images.}
%     \label{fig:14}
% \end{figure}
\subsubsection{Results on Low Quality Images in Unconstrained Environment}
\begin{figure}
\vspace{-2mm}
\subfloat[Traffic sign recognition on partially occluded and damaged images.]{\includegraphics[width=5.3cm]{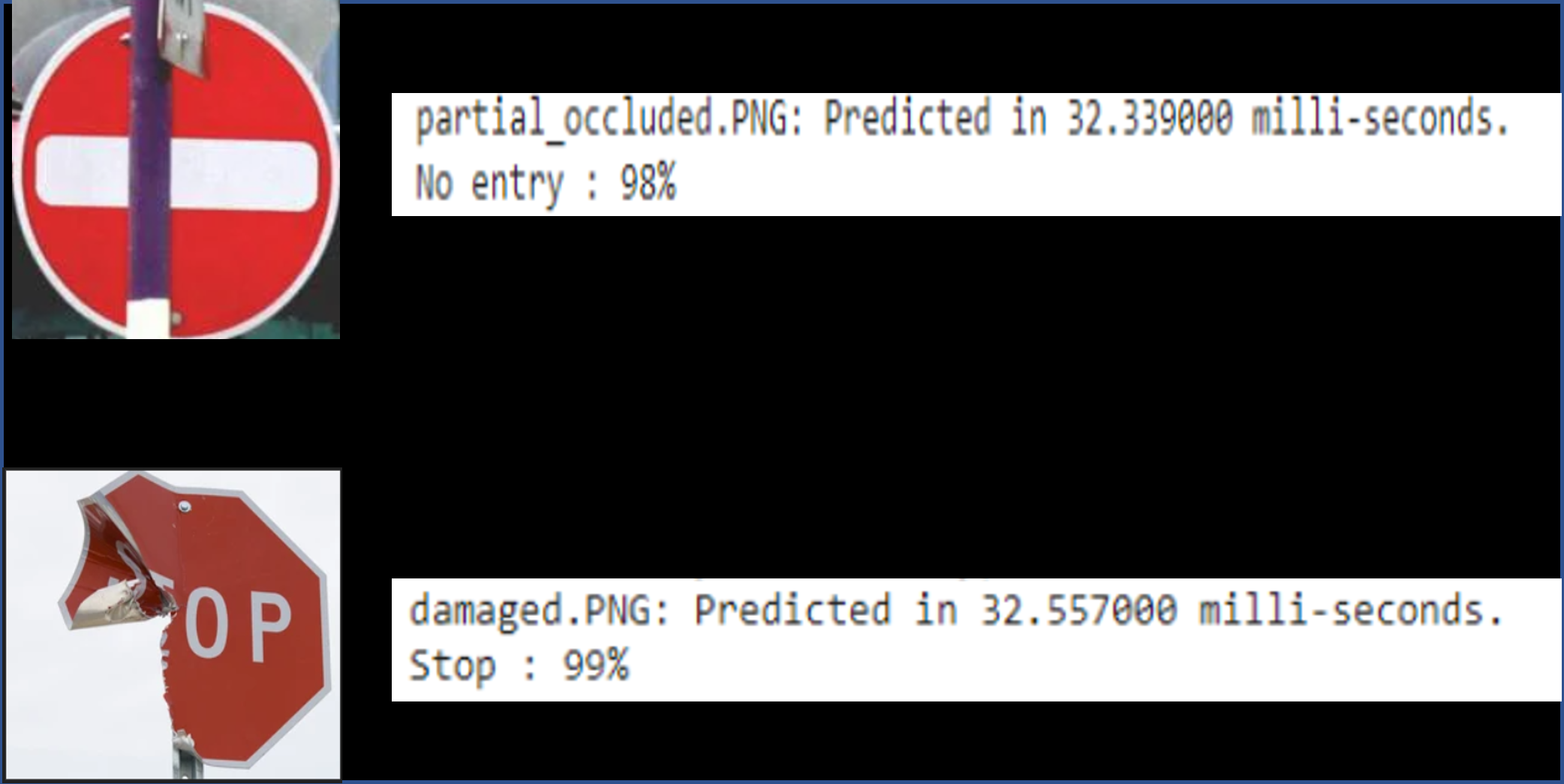}
% \centering
\label{fig:15a}}
\hfill
\subfloat[Traffic sign recognition on partially and fully illuminated images.]{\includegraphics[width=5.3cm]{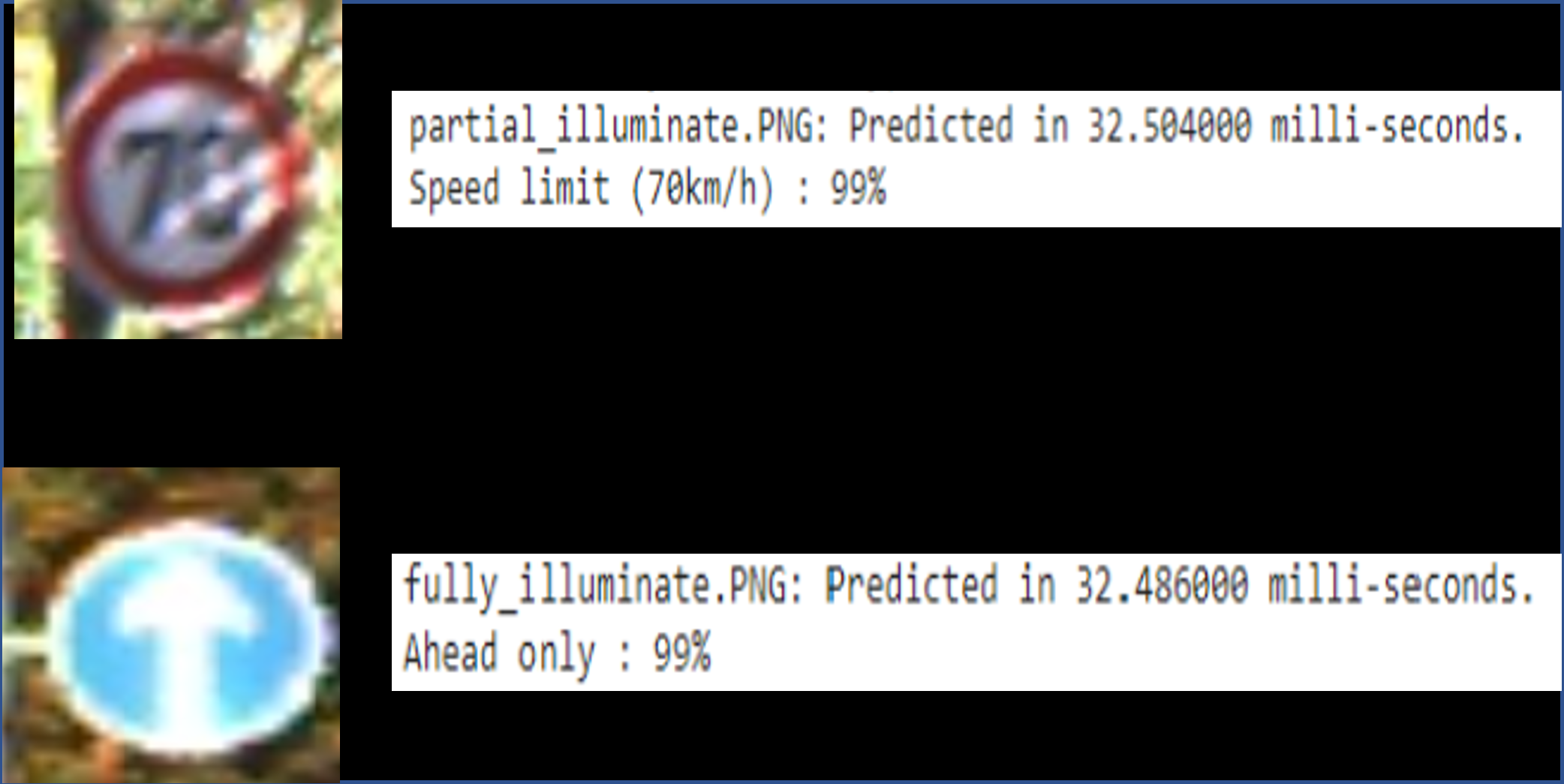}
\label{fig:15b}}
\\
\subfloat[Traffic sign recognition on faded and snow-covered images.]{\includegraphics[width=5.3cm]{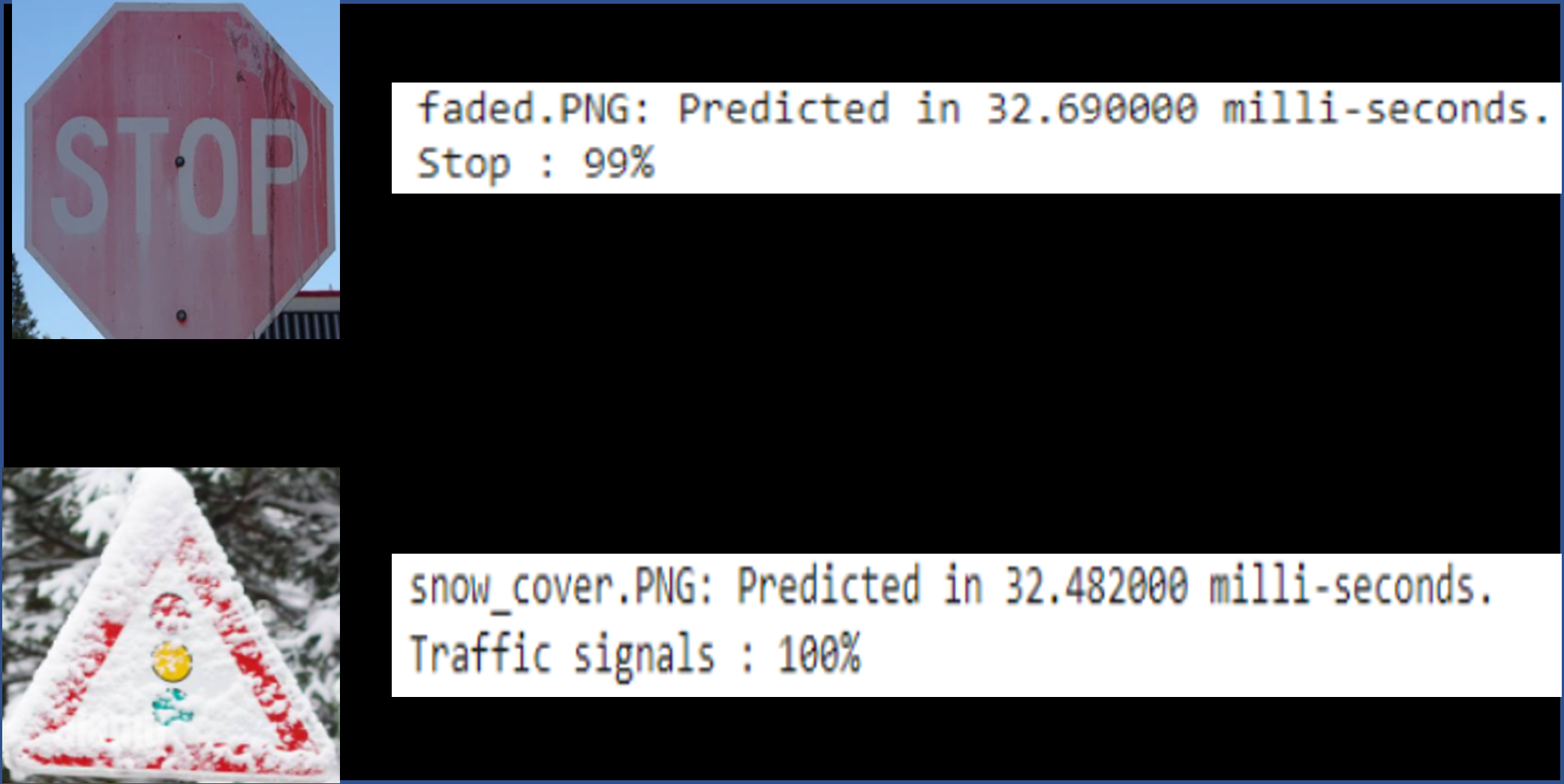}
% \centering
\label{fig:15c}}
\hfill
\subfloat[Traffic sign recognition on multiple sign images.]{\includegraphics[width=5.3cm]{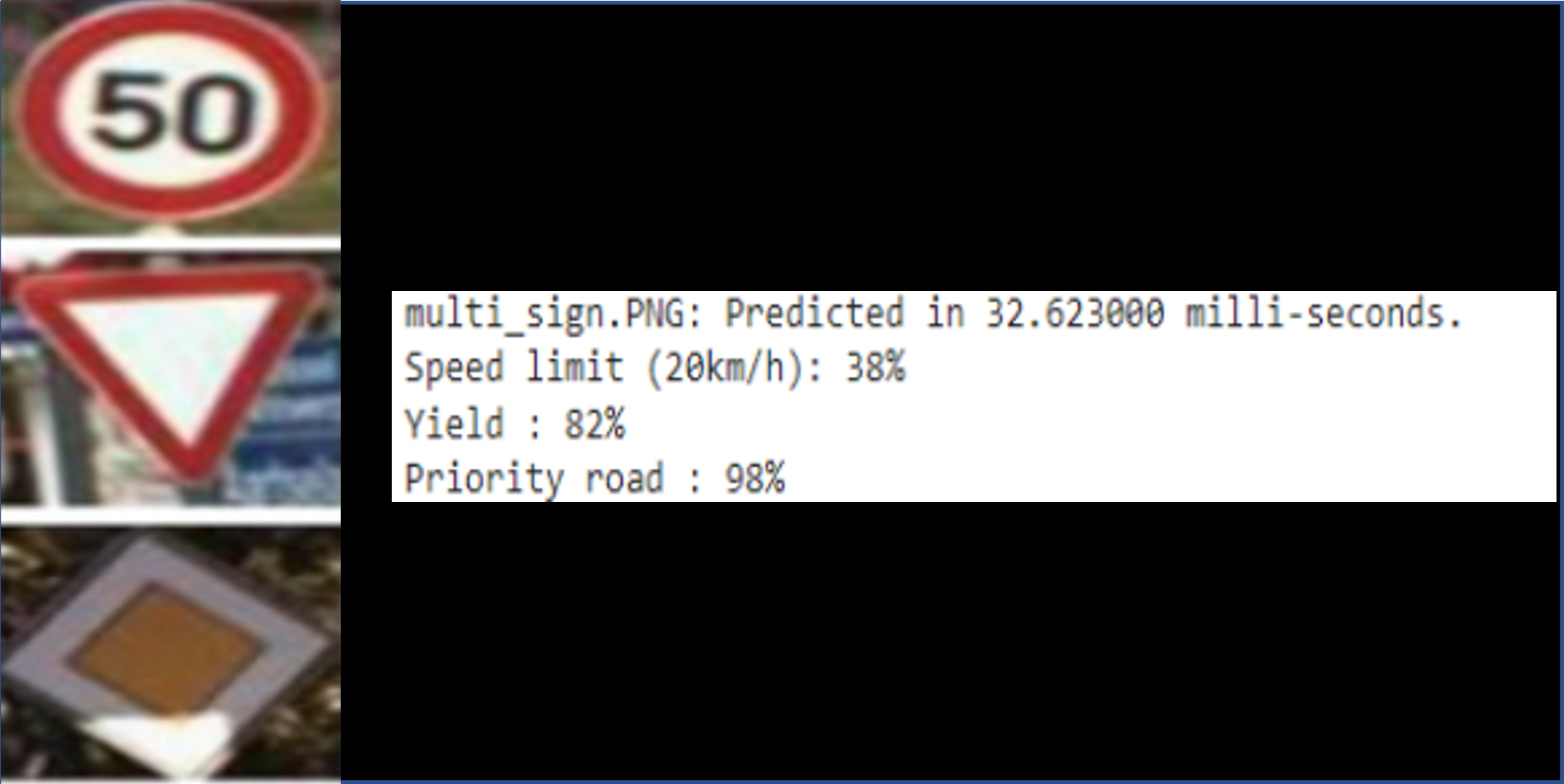}
\label{fig:15d}}
\caption{Traffic sign recognition for low quality test images.}
\label{fig:15}
\vspace{-3mm}
\end{figure}

In an unconstrained environment, it is very difficult to predict traffic signs due to partial occlusion, faded or damaged signboards, partial or full illumination, snow-covered, multiple signs in a single image, etc. The proposed approach was tested for low quality images caused by the above issues. Fig.~\ref{fig:15a} shows that the proposed model correctly predicts the traffic sign from partially occluded and damaged images. In partial or over-illuminated conditions, the proposed model correctly predicted traffic signs with high probability, as shown in Fig.~\ref{fig:15b}. Fig.~\ref{fig:15c} shows the correct prediction of the model if signs are faded or covered by snow with the probability of 99\%. With multiple traffic signs in a single image, the proposed model can predict ‘Yield’ and ‘Priority Road’ traffic signs with high probability, as shown in Fig.~\ref{fig:15d}. However, the model mispredicts the ‘Speed limit 50kmph’ sign with low confidence. From Fig.~\ref{fig:15}, it is evident that the proposed model performed well for low quality images.
\vspace{-3mm}
\subsubsection{Results on the Modified MIRNet Enhanced Images}
\begin{figure}
\vspace{-3mm}
\centering
\subfloat[Without modified MIRNet enhancement]{\includegraphics[width=5.8cm]{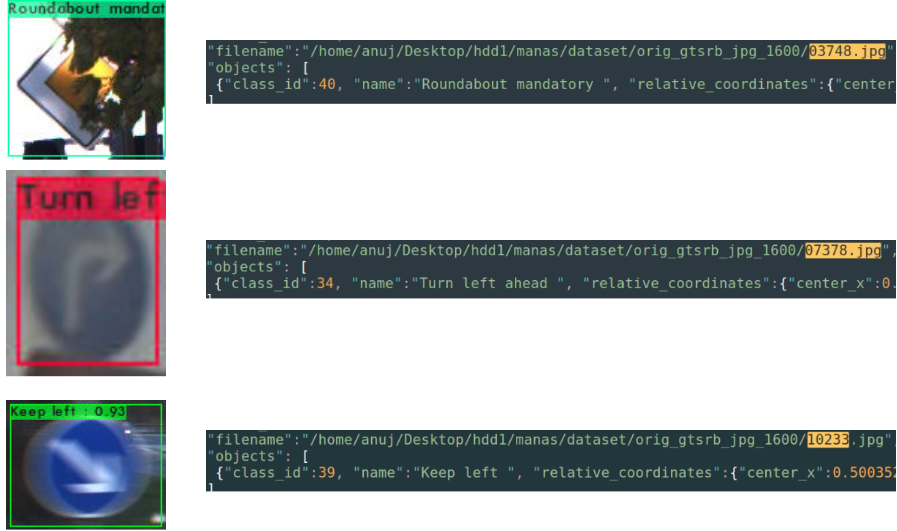}
\centering
\label{fig:16a}}
\hfill
\centering
\subfloat[With modified MIRNet enhancement of (a)]{\includegraphics[width=5.8cm]{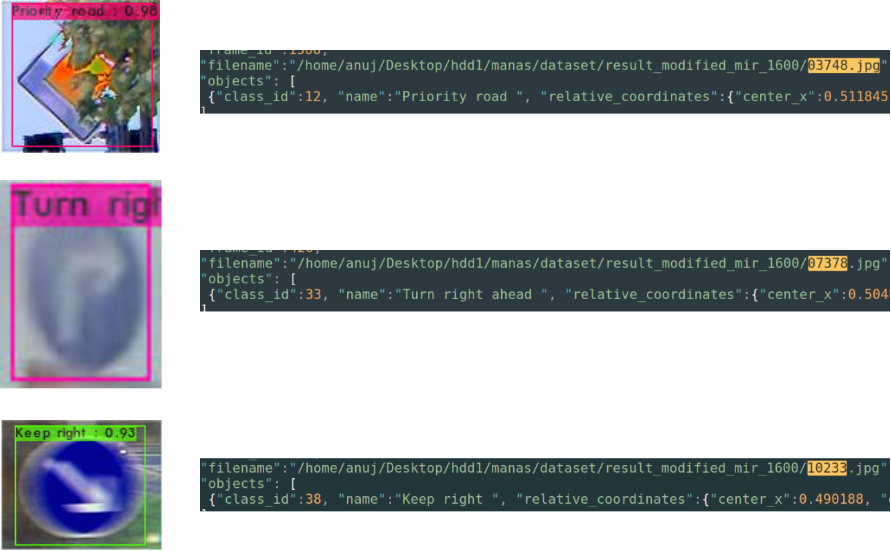}
\label{fig:16b}}
\vspace{-2mm}
\centering
\caption{Yolov4~\cite{17} prediction results with and without modified MIRNet enhancement}
\label{fig:16}
\vspace{-3mm}
\end{figure}

1600 low quality images were manually chosen and provided to the modified MIRNet~\cite{18} model. The resultant enhanced quality images from the modified MIRNet~\cite{18} model were then passed through the Yolov4~\cite{17} model. Fig.~\ref{fig:16} shows the predicted class on the original low quality images (Fig.~\ref{fig:16a}) and enhanced images (Fig.~\ref{fig:16b}). It has been observed that the predictions made by the Yolov4 model on the enhanced images are correct and uniquely localized.
% , i.e., there is a single prediction made for the traffic sign, but Yolov4 produced two predictions for the same image result on low-quality images.
% as shown in Fig.~\ref{fig:16a}. It also shows the predicted class of the corresponding enhanced quality image as shown in Fig.~\ref{fig:16b} here the predictions are made by the Yolov4 model. The prediction results are also shown in tabular format.
% (Table~\ref{tab:tab2}). 
% The actual label, predicted label of the GTSRB~\cite{2} test image with Yolov4 and predicted label of Modified MIRNet~\cite{18} enhanced image with Yolov4~\cite{17} is given in Table~\ref{tab:tab2}. 
% \begin{table*}
% % \vspace*{12pt}
% \begin{center}
% \caption{Prediction results of GTSRB~\cite{2}, Modified MIRNet~\cite{18} with Yolov4~\cite{17}}
%  \begin{tabular}{|c|c|c|c|} 
%  \hline
%  \textbf{Test Image} & \textbf{Actual Label} & \textbf{GTSRB-Yolo Pred} & \textbf{Modified MIRNET-Yolov4 Pred} \\
%  \hline
%  03748.jpg & Priority road & Roundabout mandatory & Priority road \\
%  \hline
%  07378.jpg & Turn right ahead & Turn left ahead & Turn right ahead \\
%  \hline
%  10233.jpg & Keep right & Keep left & Keep right \\ 
%  \hline
%  \hline
% \end{tabular}
% \label{tab:tab2}
% \end{center}
% \end{table*}
\vspace{-3mm}
\subsection{Results on GTSDB with Yolov4}
With the broader categories of GTSDB~\cite{1219} dataset as shown in Table~\ref{tab:tab3}, the Yolov4~\cite{17} model was trained with a lesser number of full-sized traffic sign images, which are suitable for the real-time scenarios. An average precision of 99.99\%  with Prohibitory traffic signs, 100\% with Mandatory class, 100\% with Danger signs, and 100\% with other classes has been achieved. The overall Precision of 99\%, Recall of 99\%, and the \textbf{mAP@0.5} of 100\% have been achieved for four broad categories.
\begin{table}
% \vspace{-3mm}
\caption{Testing results of GTSDB~\cite{1219} dataset: \textbf{mAP: $99.99$\%(or$~100$\%), precision=$99$\%, recall\% and F1-score=$99.99$\% }. }
\vspace{2mm}
\centering
\begin{tabular}{ | c | c| c | c| c |}
\hline
\textbf{Class} & \textbf{AP} & \textbf{TP} & \textbf{FP} \\
\hline
 Prohibitory& 99.99\% & 122 & 1
 \\
\hline
 Mandatory&100\% & 39 & 0
 \\
\hline
 Danger&100\% & 38 & 1
 \\
\hline
 Other&100\% & 39 & 0\\
\hline
\hline
\end{tabular}
\label{tab:tab3}
\vspace{-5mm}
\end{table}
\vspace{-3mm}
\subsection{Comparison of Yolov4 and Yolov4 + Modified MIRNet Results}
The following subsections provide a comprehensive evaluation of the performance of the Yolov4 model, both before and after the implemented enhancements. 
% is needed to know the actual improvement made by the model. In this subsection, we are trying to show the performance comparison before and after the Modified MIRNet enhancement. The result comparison of low-quality and enhanced-quality images
\vspace{-3mm}
\subsubsection{Comparison with Low Quality Images}
1600 low quality images were manually selected and enhanced using a modified version of MIRNet. The enhanced images were then passed through Yolov4 for prediction. When GTSRB~\cite{2} low quality images were passed through Yolov4, the \textbf{mAP} was 84.78\%. However, when the same images were enhanced through modified MIRNet~\cite{18} and passed through Yolov4, the \textbf{mAP} was increased to 90.18\%. The Yolov4~\cite{17} has demonstrated superior performance when combined with the modified MIRNet~\cite{18} model (Fig.~\ref{proposed}). 
\vspace{-3mm}
\subsubsection{Comparison with Full Dataset}
The model has originally trained with GTSRB~\cite{2} dataset. The model's performance was first checked without applying any enhancement to the test images of the full GTSRB~\cite{2} dataset. Then the low quality images were enhanced and replaced the 1600 modified MIRNet enhanced images in the GTSRB~\cite{2} test set. In both cases, the precision (accuracy) was improved by 1\% and the overall mAP@0.5 was increased from 96.65\% to 96.75\%. Thus the modified MIRNet model is able to achieve an overall improvement in mAP.
\subsection{Comparison with the Existing Work}
This subsection covers the comparative analysis of results achieved by our model on accuracy and mAP values with other existing models.
\vspace{-3mm}
\subsubsection{Comparison between Different Techniques on GTSRB Dataset}
Table~\ref{tab:tab6} shows the comparisons of results of the proposed approach with different CNN-based approaches in terms of accuracy and with earlier Yolo versions in terms of mAP. The proposed method is able to generate mAP of 96.75\%, which is better than all earlier Yolo versions. 
% On the low contrast, noisy and blurry test set images, the accuracy is increased by 5\% and the overall mAP increased from 84.78\% to 90.18\% for our model. After Modified MIRNet enhancement, the overall GTSRB~\cite{2} test-set accuracy increased by 1\% and the mAP increased from 96.65\% to 96.75\%. 
% Hence, by following the two step approach we are able to get better mAP and accuracy than the other techniques shown Table~\ref{tab:tab6}. The accuracy achieved by our model is better than CNN technique used in~\cite{14}, it is comparable to the CNN technique used in~\cite{13}. In terms of mAP@0.5 score, our approach easily beats Yolov3~\cite{16,8}, Yolov3 Tiny~\cite{16,8}, and Compact Squeezed Yolov4~\cite{16}.
\begin{table}
% \vspace*{12pt}
\vspace{-4mm}
\caption{Results Comparison of Different Models with GTSRB}
\centering
\begin{tabular}{ | p{5cm} | c | c | c | }
\hline
\textbf{Algorithm} & \textbf{mAP \% @ 0.50} & \textbf{Av. IOU} & \textbf{Accuracy}\\
\hline
CNN~\cite{14} & - & - & 95\%
\\
\hline
CNN (NVIDIA Jetson Embedded TX1 System)~\cite{15} & - & - & 96.2\%
\\
\hline
Yolov3~\cite{16,8} & 94.05 & 75.63 & -
 \\
\hline
Yolov3 Tiny~\cite{16,8} & 90.04 & 75.26 & -
 \\
\hline
Yolov4 Tiny~\cite{16,17} & 97.06 & 74.15 & -
\\
\hline
Compact New Squeezed Yolo\cite{16} & 95.44 & 71.68 & -
\\
\hline
\textbf{Yolov4 without enhancement} & \textbf{96.65} & \textbf{78.24} & \textbf{95\%}
\\
\hline
\textbf{modified MIRNet+Yolov4 (proposed)} & \textbf{96.75} & \textbf{78.92} & \textbf{96\%}
\\
\hline
\hline
\end{tabular}
\label{tab:tab6}
\vspace{-5mm}
\end{table}
\vspace{-3mm}
\subsubsection{Comparison between Different Techniques Trained on GTSDB Dataset}
The Yolov4 model was trained with the GTSDB dataset for the four broad categories. A precision of 99\%, recall of 99\% and mAP of 100\% with the Yolov4~\cite{17} model was achieved. Table~\ref{tab:tab7} shows the comparisons of the different traffic sign detector results with Yolov4~\cite{17}.
\begin{table}
\vspace{-5mm}
\small
\caption{Results comparison of different models with GTSDB}
\centering
\begin{tabular}{ | c | c | c |  }
\hline
\textbf{Detection algorithms}&\textbf{Backbone networks in algorithms}&\textbf{mAP (\%)}\\
\hline
Faster R-CNN~\cite{5,6} & ResNet50	& 97.9
\\
\hline
SSD~\cite{5,7}	& VGG16 & 93.2
\\
\hline
Yolov3~\cite{5,8} & DarkNet53 & 93.8
\\
\hline
SSD-RP\cite{5} & VGG16 & 95.4
\\
\hline
RetinaNet\cite{5} & ResNet50 & 96.7
\\
\hline
\textbf{Yolov4}& \textbf{CSPDarknet53~\cite{23}} & \textbf{99.99}
\\
\hline
\hline
\end{tabular}
\label{tab:tab7}
\vspace{-4mm}
\end{table}
As mentioned in Table~\ref{tab:tab7}, mAP achieved by Faster R-CNN~\cite{5,3} is 97.9\% using ResNet50 backbone, SSD~\cite{5,7} is 93.2\% using VGG16 backbone, Yolov3~\cite{5,8} is 93.8\% using Darknet53 backbone, SSD-RP~\cite{5} is 95.4\% using ResNet50 backbone, and RetinaNet~\cite{5} is 96.7\%. All the mAP percentages are lesser than Yolov4~\cite{17} using CSPDarknet53~\cite{23} as a backbone.

\section{Conclusion}
\label{sec:five}
This paper proposed a two-step approach for traffic sign recognition. Wherein the first step, a Modified MIRNet was used for low quality image enhancement, and in the second step, the Yolov4 model was employed to produce better output results in terms of mAP and accuracy. The experiments were performed on the GTSDB and GTSRB datasets. Through experiments, the proposed approach has achieved higher mAP and accuracy values with both broader categories (4 categories) and fine-grained classes (43). The modified MIRNet model has also achieved a higher mAP value on low quality images. In future works, more changes can be made to the proposed architecture using other enhancement techniques. The changes can be made to the PAN in the Neck part of Yolov4. Also, other state-of-the-art algorithms can be tested with the GTSRB dataset to check the model performance.
\bibliographystyle{splncs04}
\bibliography{samplepaper}

\begin{thebibliography}{10}
\providecommand{\url}[1]{\texttt{#1}}
\providecommand{\urlprefix}{URL }
\providecommand{\doi}[1]{https://doi.org/#1}

\bibitem{17}
Bochkovskiy, A., Wang, C.Y., Liao, H.Y.M.: Yolov4: Optimal speed and accuracy
  of object detection. arXiv preprint arXiv:2004.10934  (2020)

\bibitem{8}
Farhadi, A., Redmon, J.: Yolov3: An incremental improvement. In: Computer
  Vision and Pattern Recognition (CVPR). vol.~1804, pp.~1--6. Springer
  Berlin/Heidelberg, Germany (2018)

\bibitem{21}
Girshick, R.: Fast r-cnn. In: Proceedings of the IEEE international conference
  on computer vision. pp. 1440--1448 (2015)

\bibitem{15}
Han, Y., Oruklu, E.: Traffic sign recognition based on the nvidia jetson tx1
  embedded system using convolutional neural networks. In: 2017 IEEE 60th
  International Midwest Symposium on Circuits and Systems (MWSCAS). pp.
  184--187. IEEE (2017)

\bibitem{24}
He, K., Zhang, X., Ren, S., Sun, J.: Spatial pyramid pooling in deep
  convolutional networks for visual recognition. IEEE Transactions on Pattern
  Analysis and Machine Intelligence  \textbf{37}(9),  1904--1916 (2015)

\bibitem{1219}
Houben, S., Stallkamp, J., Salmen, J., Schlipsing, M., Igel, C.: Detection of
  traffic signs in real-world images: The german traffic sign detection
  benchmark. In: The 2013 International Joint Conference on Neural Networks
  (IJCNN). pp.~1--8. Ieee (2013)

\bibitem{1}
Karthika, R., Parameswaran, L.: A novel convolutional neural network based
  architecture for object detection and recognition with an application to
  traffic sign recognition from road scenes. Pattern Recognition and Image
  Analysis  \textbf{32}(2),  351--362 (2022)

\bibitem{16}
Khnissi, K., Jabeur, C.B., Seddik, H.: Implementation of a compact traffic
  signs recognition system using a new squeezed yolo. International Journal of
  Intelligent Transportation Systems Research  \textbf{20}(2),  466--482 (2022)

\bibitem{3}
Lin, T.Y., Doll{\'a}r, P., Girshick, R., He, K., Hariharan, B., Belongie, S.:
  Feature pyramid networks for object detection. In: Proceedings of the IEEE
  Conference on Computer Vision and Pattern Recognition (CVPR). pp. 2117--2125
  (2017)

\bibitem{25}
Liu, S., Qi, L., Qin, H., Shi, J., Jia, J.: Path aggregation network for
  instance segmentation. In: Proceedings of the IEEE conference on computer
  vision and pattern recognition. pp. 8759--8768 (2018)

\bibitem{7}
Liu, W., Anguelov, D., Erhan, D., Szegedy, C., Reed, S., Fu, C.Y., Berg, A.C.:
  Ssd: Single shot multibox detector. In: Computer Vision--ECCV 2016: 14th
  European Conference, Amsterdam, The Netherlands, October 11--14, 2016,
  Proceedings, Part I 14. pp. 21--37. Springer (2016)

\bibitem{4}
Novak, B., Ili{\'c}, V., Pavkovi{\'c}, B.: Yolov3 algorithm with additional
  convolutional neural network trained for traffic sign recognition. In: 2020
  Zooming Innovation in Consumer Technologies Conference (ZINC). pp. 165--168.
  IEEE (2020)

\bibitem{14}
Patil, D., Poojari, A., Choudhary, J., Gaglani, S.: Cnn based traffic sign
  detection and recognition on real time video. International Journal of
  Engineering Research \& Technology (IJERT)  \textbf{9}(3), ~1--5 (2021)

\bibitem{22}
Redmon, J., Divvala, S., Girshick, R., Farhadi, A.: You only look once:
  Unified, real-time object detection. In: Proceedings of the IEEE conference
  on computer vision and pattern recognition. pp. 779--788 (2016)

\bibitem{6}
Ren, S., He, K., Girshick, R., Sun, J.: Faster r-cnn: Towards real-time object
  detection with region proposal networks. IEEE Transactions on Pattern
  Analysis and Machine Intelligence  \textbf{39}(6),  1137--1149 (2017)

\bibitem{2}
Stallkamp, J., Schlipsing, M., Salmen, J., Igel, C.: Man vs. computer:
  Benchmarking machine learning algorithms for traffic sign recognition. Neural
  Networks  \textbf{32},  323--332 (2012)

\bibitem{13}
Sun, Y., Ge, P., Liu, D.: Traffic sign detection and recognition based on
  convolutional neural network. In: 2019 Chinese Automation Congress (CAC). pp.
  2851--2854. IEEE (2019)

\bibitem{9}
Tai, Y.W., Jia, J., Tang, C.K.: Soft color segmentation and its applications.
  IEEE Transactions on Pattern Analysis and Machine Intelligence
  \textbf{29}(9),  1520--1537 (2007)

\bibitem{23}
Wang, C.Y., Liao, H.Y.M., Wu, Y.H., Chen, P.Y., Hsieh, J.W., Yeh, I.H.: Cspnet:
  A new backbone that can enhance learning capability of cnn. In: Proceedings
  of the IEEE/CVF conference on computer vision and pattern recognition
  workshops. pp. 390--391 (2020)

\bibitem{10}
Wang, Q., Liu, X.: Traffic sign segmentation in natural scenes based on color
  and shape features. In: 2014 IEEE Workshop on Advanced Research and
  Technology in Industry Applications (WARTIA). pp. 374--377. IEEE (2014)

\bibitem{27}
Wang, Z., Wang, J., Li, Y., Wang, S.: Traffic sign recognition with lightweight
  two-stage model in complex scenes. IEEE Transactions on Intelligent
  Transportation Systems  \textbf{23}(2),  1121--1131 (2020)

\bibitem{26}
Wei, C., Wang, W., Yang, W., Liu, J.: Deep retinex decomposition for low-light
  enhancement. arXiv preprint arXiv:1808.04560  (2018)

\bibitem{5}
Wu, J., Liao, S.: Traffic sign detection based on ssd combined with receptive
  field module and path aggregation network. Computational Intelligence and
  Neuroscience  \textbf{2022} (2022)

\bibitem{20}
Yang, Y., Luo, H., Xu, H., Wu, F.: Towards real-time traffic sign detection and
  classification. IEEE Transactions on Intelligent Transportation Systems
  \textbf{17}(7),  2022--2031 (2015)

\bibitem{11}
Yao, C., Wu, F., Chen, H.j., Hao, X.l., Shen, Y.: Traffic sign recognition
  using hog-svm and grid search. In: 2014 12th International Conference on
  Signal Processing (ICSP). pp. 962--965. IEEE (2014)

\bibitem{18}
Zamir, S.W., Arora, A., Khan, S., Hayat, M., Khan, F.S., Yang, M.H., Shao, L.:
  Learning enriched features for real image restoration and enhancement. In:
  Computer Vision--ECCV 2020: 16th European Conference, Glasgow, UK, August
  23--28, 2020, Proceedings, Part XXV 16. pp. 492--511. Springer (2020)

\end{thebibliography}
% LNCS Homepage, \url{http://www.springer.com/lncs}. Last accessed 4
% \end{thebibliography}
\end{document}